\documentclass[11pt]{article}

\usepackage[preprint]{acl}
\usepackage{times}
\usepackage{latexsym}
\usepackage{float}
\usepackage[most]{tcolorbox}
\usepackage{enumitem}
\usepackage[T1]{fontenc}
\usepackage[utf8]{inputenc}
\usepackage{microtype}
\usepackage{inconsolata}
\usepackage{graphicx}
\usepackage{tabularx}
\usepackage{nicematrix,booktabs}
\usepackage[most]{tcolorbox}
\usepackage{fvextra} \DefineVerbatimEnvironment{PromptVerbatim}{Verbatim}{
  fontsize=\small,
  breaklines=true,
  breakanywhere=true,
  commandchars=\\\{\}
}
\newtcolorbox{promptbox}[1]{
  enhanced,
  colback=white,
  colframe=black,
  boxrule=0.5pt,
  arc=2pt,
  left=6pt,right=6pt,top=6pt,bottom=6pt,
  title=\textbf{#1},
  fonttitle=\small,
}

\title{Concept Tokens: Learning Behavioral Embeddings Through Concept Definitions}

\author{Ignacio Sastre \and Aiala Rosá \\
Instituto de Computación, Facultad de Ingeniería, Universidad de la República \\ Montevideo, Uruguay\\\texttt{\{isastre,aialar\}@fing.edu.uy}}

\begin{document}
\maketitle
\begin{abstract}
We propose \emph{Concept Tokens}, a lightweight method that adds a new special token to a pretrained LLM and learns only its embedding from multiple natural language definitions of a target concept, where occurrences of the concept are replaced by the new token.
The LLM is kept frozen and the embedding is optimized with the standard language-modeling objective.
We evaluate Concept Tokens in three settings.
First, we study \emph{hallucinations} in closed-book question answering on HotpotQA and find a directional effect: negating the hallucination token reduces hallucinated answers mainly by increasing abstentions, whereas asserting it increases hallucinations and lowers precision.
Second, we induce \emph{recasting}, a pedagogical feedback strategy for second language teaching, and observe the same directional effect.
Moreover, compared to providing the full definitional corpus in-context, concept tokens better preserve compliance with other instructions (e.g., asking follow-up questions).
Finally, we include a qualitative study with the Eiffel Tower and a fictional ``Austral Tower'' to illustrate what information the learned embeddings capture and where their limitations emerge.
Overall, Concept Tokens provide a compact control signal learned from definitions that can steer behavior in frozen LLMs.\footnote{Code and data are available at: \url{https://github.com/nsuruguay05/concept_tokens}}
\end{abstract}

\section{Introduction}
\label{sec:introduction}

Recent work has highlighted the importance of the input embedding layer in large language models (LLMs), showing that it is possible to learn particular representations in this space that induce specific generation behaviors~\cite{sastre-rosa-2025-memory, kuratov-etal-2025-cramming}.
This motivates the search for structured, interpretable representations in the embedding space beyond the original vocabulary.

Most vocabulary tokens in modern LLMs are subword units, and even full-word tokens can be polysemous.
In both cases, individual tokens rarely correspond to well-formed concepts.
Instead, concept-level meaning emerges through contextual processing across Transformer layers, as the model builds contextual representations for each token.

Humans can learn by conceptualizing: we can often acquire new concepts from language alone, for instance, by reading a definition, with few or no task examples.
Motivated by this, we ask whether an LLM can also internalize a concept from \textit{definition-only supervision}: instead of training on labeled examples, we use the concept's natural language definition. 

We introduce \emph{Concept Tokens}: special tokens whose embeddings are optimized using only definitions of a target concept, while keeping the pretrained model frozen.
This enables (1) adding previously unknown concepts to a pretrained LLM and (2) inducing behavior changes that are naturally associated with understanding a concept, without training on behavioral examples.

We focus on relatively small models, motivated by educational applications that often benefit from on-premise deployment to ensure data privacy and accessibility across different contexts (e.g., rural schools).
A compact representation of a concept that can steer the model's behavior is especially useful for smaller models, which may struggle to reliably follow lengthy or nuanced instructions from context alone.

We evaluate Concept Tokens in three settings:
(1) steering the model away from \emph{hallucinations} in closed-book question answering;
(2) inducing the \emph{recasting} feedback strategy used in second language teaching;
and (3) a qualitative analysis using the \emph{Eiffel Tower} and a fictional \emph{``Austral Tower''}.
Together, these experiments illustrate both the capabilities and limitations of concept tokens as compact behavioral representations.

The remainder of the paper is organized as follows: Section~\ref{sec:related-work} reviews related work; Section~\ref{sec:concept-tokens} introduces Concept Tokens and our central hypothesis; Section~\ref{sec:experiments} describes the three experiments; and Section~\ref{sec:conclusions} concludes with limitations and future directions.

\section{Related work}
\label{sec:related-work}

Soft prompting adapts a frozen language model by learning a small set of continuous prompt parameters (special tokens) that condition the model's behavior without updating its weights.
P*-tuning~\cite{li-liang-2021-prefix,lester-etal-2021-power,liu-etal-2022-p} refers to a family of soft prompting methods that aim at optimizing a set of vectors and using them as a prefix for a specific task.
A related line of work studies prompt compression, aiming to replace long natural-language contexts with compact learned representations (soft tokens) to reduce inference cost while retaining task-relevant information~\cite{li-etal-2025-prompt,li-etal-2025-500xcompressor}.

Recently, concurrent work has shown that it is possible to compress large sequences of text into a single embedding (soft token) without information loss, effectively constructing reversible sentence embeddings~\cite{sastre-rosa-2025-memory, kuratov-etal-2025-cramming}.
By maintaining a pretrained LLM frozen and repeatedly optimizing a single input vector using a language modeling loss, effectively overfitting on the target sequence, when using the optimized embedding as input to the frozen LLM, the model reconstructs the original sequence token by token.
This demonstrates how big of an impact the input embeddings have on the model, and introduces a mechanism to learn particular representations in the embedding space that induce specific generation behaviors, which inspired the present work.

Another line of research is Large Concept Models (LCMs)~\cite{lcmteam2024largeconceptmodelslanguage} which consists of shifting language modeling from token sequences to a higher-level semantic space.
Here, a concept corresponds to a sentence embedding, which is slightly different to the notion of concept we use in this work.
Concretely, the input is segmented into sentences, which are then encoded with a sentence embedding encoder, then processed by an LCM to generate a new sequence of concepts, which are decoded by a sentence embedding decoder into sequences of tokens.
Concept tokens, as we present them in this work, rather than replacing token-level generation, are a lightweight way to introduce a new concept and steer behavior, while preserving the original model frozen.

Closer to our proposal are latent steering vectors~\cite{subramani-etal-2022-extracting}, a way to manipulate internal activations rather than input embeddings.
It is possible to extract vectors from a frozen LLM that, when added to its hidden states, can strongly steer generation toward a target sentence. 
Another work proposes Generation with Concept Activation Vector (GCAV)~\cite{zhang2025controllinglargelanguagemodels}, a lightweight control by learning a direction corresponding to an attribute (e.g., toxicity, sentiment, topic) and then steering generation by injecting/removing that concept direction at selected layers. 

Mechanistic interpretability work has shown that sparse autoencoders (SAEs) can extract sparse, human-interpretable features from LLM activations, with evidence that some learned features behave approximately aligned with a coherent concept and high-level behaviors can be partially attributed to a set of discoverable latent directions~\cite{bricken2023towards, templeton2024scaling}.
Recent studies leverage SAE features for steering: they select features and then intervene to improve downstream behavior via feature-guided activation edits~\cite{cho-hockenmaier-2025-toward}. 
Persona vectors~\cite{chen2025personavectorsmonitoringcontrolling} identify activation directions corresponding to behavioral traits (including hallucination propensity) and use them to monitor or modulate those traits. 
In our proposal, rather than explicitly extracting features from activations, we learn an input embedding that elicits a behavior from the frozen model by inserting it in prompts and may indirectly activate the relevant internal features.

\section{Concept Tokens}
\label{sec:concept-tokens}

\begin{figure*}[t]
  \centering
    \includegraphics[width=0.85\linewidth]{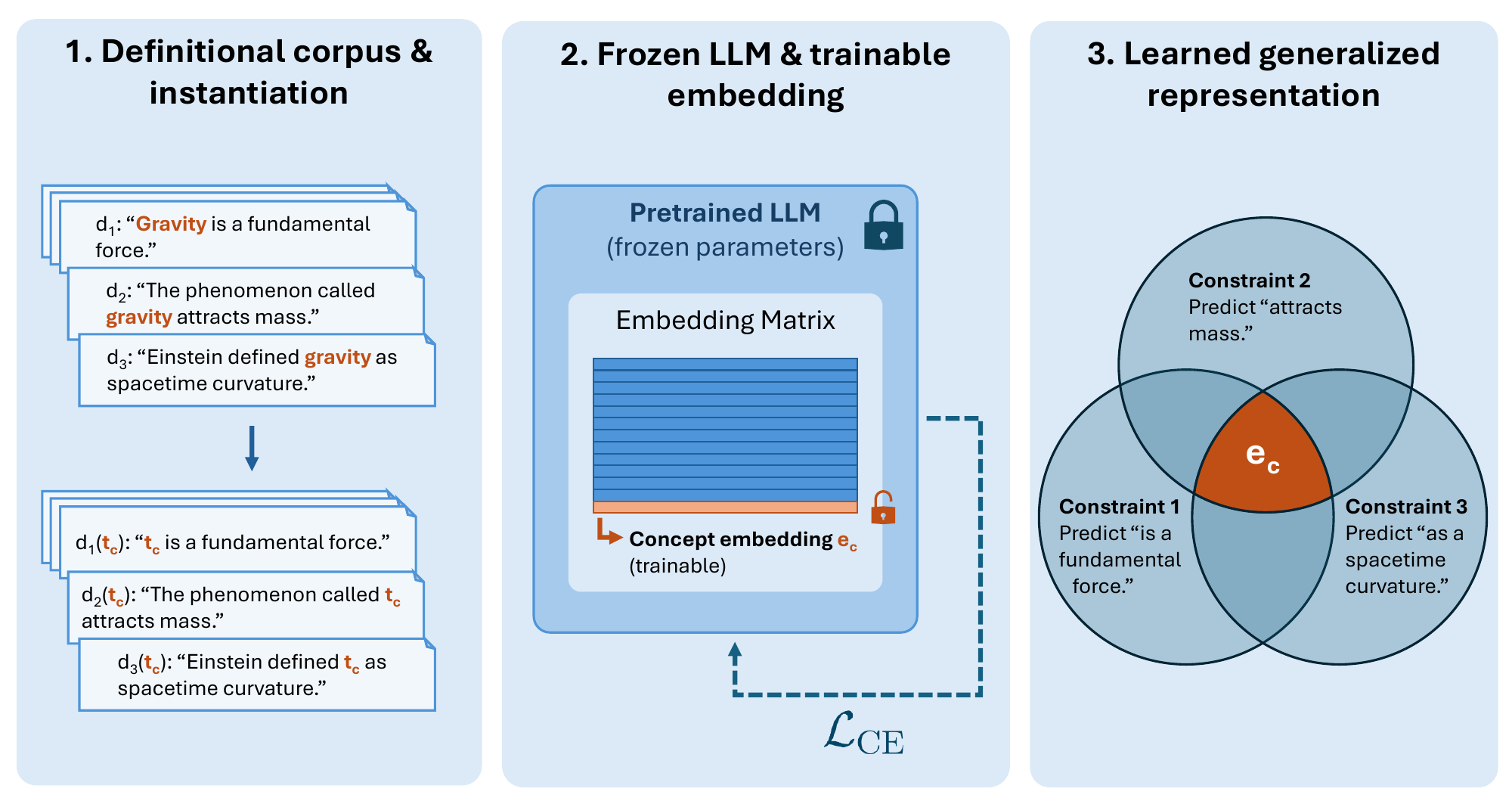}
  \caption{Overview of concept tokens. (1) We build a definitional corpus for a concept and instantiate it by replacing each mention with a new token $t_c$. (2) Keeping the LLM frozen, we optimize only the concept embedding $e_c$ by minimizing cross-entropy loss on the instantiated corpus. (3) The learned embedding captures a generalized representation that best satisfies the (potentially competing) constraints imposed by multiple definitions.}
  \label{fig:figure_ct}
\end{figure*}

A \textit{concept token} $t_c$ for a concept $c$ is a new special token added to the vocabulary of a pretrained LLM with a corresponding embedding $e_c$.

We call a \textit{definitional corpus} to a set of definitions for a concept $c$: $\mathcal{D}_c = \{d^c_1,d^c_2,...,d^c_n\}$, where each definition $d^c_i$ has at least one explicit occurrence of the concept $c$.

For each definition $d^c_i$, we replace each occurrence of the concept $c$ with the concept token $t_c$, which we note $d^c_i(t_c)$, thereby obtaining an \textit{instantiated definitional corpus} $\mathcal{D}_c(t_c) = \{d^c_1(t_c),d^c_2(t_c),...,d^c_n(t_c)\}$.

We will optimize the embedding $e_c$ using $\mathcal{D}_c(t_c)$, while maintaining the entire model frozen, including the rest of the pretrained embeddings.
The optimization process tries to minimize the cross-entropy loss for the language modeling objective (same as in pretraining).

We can think of each instantiated definition as a restriction over the concept token.
By minimizing the cross-entropy loss by only adjusting the embedding $e_c$, it needs to encode a good representation for predicting simultaneously all the definitions in the definitional corpus.

We can observe that the memory token, as presented in \cite{sastre-rosa-2025-memory, kuratov-etal-2025-cramming}, can be seen as a special case of a concept token, where the definitional corpus consists of only one definition (the text to memorize) and one occurrence of the concept at the beginning of that definition.
In this scenario, there is only one restriction imposed.
Therefore, the optimization process is capable of encoding in $e_c$ a lossless representation that can reconstruct the definition perfectly.

In the more general case with multiple definitions, there are multiple restrictions with possibly contradicting objectives, for example, if two definitions start with the concept.
In that situation, given only the concept token as input, the model is expected to generate two different sequences of tokens, corresponding to the two definitions.
Given that the objective of this work is not to memorize the definitions, but rather to learn good representations of concepts to be used as input for a pretrained LLM, this is exactly what we are looking for, as the embedding $e_c$ has to learn a generalized representation capable of obeying as best as possible each restriction given by each definition.

Therefore, we propose the following hypothesis: given a definitional corpus $\mathcal{D}_c$ with multiple definitions, the best possible embedding $e_c$ to be learned in the process described before, is the one that better captures the original concept $c$.

At inference time, the concept token $t_c$ is used like any other token: it is simply inserted into the prompt (e.g., in the system instruction or as part of a prefix in completion-style prompting).
Moreover, the token can be used directionally: asserting $t_c$ in the instruction encourages behavior associated with the concept, while negating it discourages it.

\section{Experiments}
\label{sec:experiments}

We evaluate Concept Tokens in three complementary settings.
The first two experiments study behavioral steering: (1) reducing hallucinations in closed-book question answering, and (2) inducing the recasting feedback strategy in second language teaching.
Finally, we present a qualitative study with a real concept (the Eiffel Tower) and a fictional one (the Austral Tower) to better understand what a concept token embedding can encode, how it generalizes beyond its definitional corpus, and where its limitations emerge.

All experiments use Llama 3.1 8B  Instruct~\cite{grattafiori2024llama3herdmodels} quantized to 4-bit weights, and generation is performed with greedy decoding. 
Detailed training hyperparameters and optimization settings are reported in Appendix~\ref{appendix:training-details}.

\subsection{Hallucinations}
\label{sec:experiments-hallucinations}

Hallucinations refer to the tendency of LLMs to produce fluent, seemingly plausible outputs that are not supported by facts~\cite{survey-hallucinations}.
This behavior is particularly problematic in question answering, including retrieval-augmented generation (RAG) settings.
Recent work argues that hallucinations persist in part because current training and evaluation setups reward plausible guessing over the admission of uncertainty, effectively incentivizing models to answer even when unsure to optimize benchmark performance~\cite{kalai2025languagemodelshallucinate}.

We test whether a concept token can internalize the notion of hallucinations from definitional text and be used as a compact representation to induce a behavioral shift at inference time. Additionally, we evaluate whether this mechanism can reduce hallucinated answers while preserving a high rate of correct responses.

\subsubsection{Setup}
\label{sec:experiments-hallucinations-setup}

The definitional corpus was synthetically generated with GPT-5 (OpenAI)~\footnote{\url{https://openai.com/index/introducing-gpt-5/}}.
We instructed the model to produce multiple paragraphs, each providing a distinct (possibly redundant) definition of the hallucinations concept, while ensuring that every occurrence of the term appears explicitly as \emph{hallucinations}.
The resulting corpus contains 20 paragraphs and 102 occurrences of \emph{hallucinations} (see Appendix~\ref{appendix:hallucinations-definitions} for representative excerpts).
We replace each occurrence with the concept token to train its embedding.

The task we use to evaluate the concept token is \emph{closed-book} generative question answering. 
That is, given a question and no external information, the model must generate an answer.
In our analysis, we distinguish three outcomes: the model answers correctly, produces a hallucinated answer, or abstains from answering.

To carry out this evaluation, we use HotpotQA~\cite{yang-etal-2018-hotpotqa}, a large-scale dataset of multi-hop questions derived from Wikipedia articles.
Each instance includes a question, a gold answer, and supporting context (paragraphs and supporting facts).
We evaluate on a subset of 1000 instances from the validation set (out of 7405 total instances).
The questions vary in complexity, ranging from relatively concrete, factoid questions (Figure~\ref{fig:examples-hotpotqa}, Example~2) to more compositional or less direct questions (Figure~\ref{fig:examples-hotpotqa}, Example~1).

We provide only the question to the model (discarding the context) and use an LLM-as-a-judge framework~\cite{zheng2023judging} to classify each generated response as \textbf{(1)} \emph{Correct}, \textbf{(2)} \emph{No Answer}, or \textbf{(3)} \emph{Hallucination}.
We use Gemini~2.5 Flash as the judge.
The prompt defines each category and asks the model to assign a label given the question, the generated answer, and the gold answer.
The full prompt is provided in Appendix~\ref{appendix:prompt-llm-judge}.
To validate this evaluation procedure, we measure agreement between the judge and human annotations on 100 instances, obtaining Cohen's $\kappa$ = 0.88.

To determine how well the concept token is capturing the \emph{hallucinations} concept, we evaluate three different settings that differ only in the system instruction (all other settings are held fixed):

\begin{enumerate}
    \item \textbf{Concept token negated.} The system prompt states: \textit{``You are a helpful assistant. Do not generate $t_c$.''}, where $t_c$ denotes the concept token trained on the definitional corpus for \emph{hallucinations}. We negate it to encourage abstention rather than guessing.
    
    \item \textbf{No instruction.} The system prompt is simply: \textit{``You are a helpful assistant''}.
    
    \item \textbf{Concept token asserted.} The system prompt states: \textit{``You are a helpful assistant. Generate $t_c$.''}. This is identical to (1) except that the negation is removed. We expect this change to increase hallucinations and reduce correct answers.
\end{enumerate}

Additionally, we define two prompting baselines that don't use the concept token, for comparison:

\begin{enumerate}
    \item \textbf{Hallucinations mention.} Same as \emph{Concept token negated}, but instead of using the concept token we use the word \emph{hallucinations} (i.e., \textit{``Do not generate hallucinations.''}).
    
    \item \textbf{Definitional corpus in-context.} Same as (1), but we prepend the full definitional corpus to the prompt, providing the definition in-context.
\end{enumerate}

Refer to Appendix~\ref{appendix:prompts-hallucination} for the complete prompts.

We include \emph{definitional corpus in-context} as a strong baseline despite its larger prompt length, since it represents the standard alternative to concept tokens: providing the definition at inference time.

\subsubsection{Results}
\label{sec:experiments-hallucinations-results}

Table~\ref{tab:hallucinations-results} reports the LLM-as-a-judge evaluation, along with the metric \emph{Precision}, defined as the proportion of correct answers among attempted answers (i.e., excluding \emph{No Answer}).

\begin{table}[t]
\centering
\small
\setlength{\tabcolsep}{3.5pt}
\renewcommand{\arraystretch}{1.15}
\begin{tabular}{lccccc}
\toprule
\textbf{Method} & \textbf{Correct} & \textbf{Halluc.} & \textbf{No answer} & \textbf{Prec.} \\
\midrule
$t_c$ negated  & 17.60 & 21.90 & 60.50 & 44.56 \\
No instruction & 25.10 & 28.70 & 46.20 & 46.65 \\
$t_c$ asserted & 16.50 & 31.20 & 52.30 & 34.59 \\
\midrule
Explicit halluc. & 15.50 & 19.10 & 65.40 & 44.80 \\
Def. in-context & 17.50 & 21.60 & 60.90 & 44.76 \\
\bottomrule
\end{tabular}
\caption{Percentage of outputs in each category, as well as $\mathrm{Prec}=\#\mathrm{Correct}/(\#\mathrm{Correct}+\#\mathrm{Hallucination})$ (precision among attempted answers).}
\label{tab:hallucinations-results}
\end{table}

\begin{figure}[t]
  \centering
    \includegraphics[width=\linewidth]{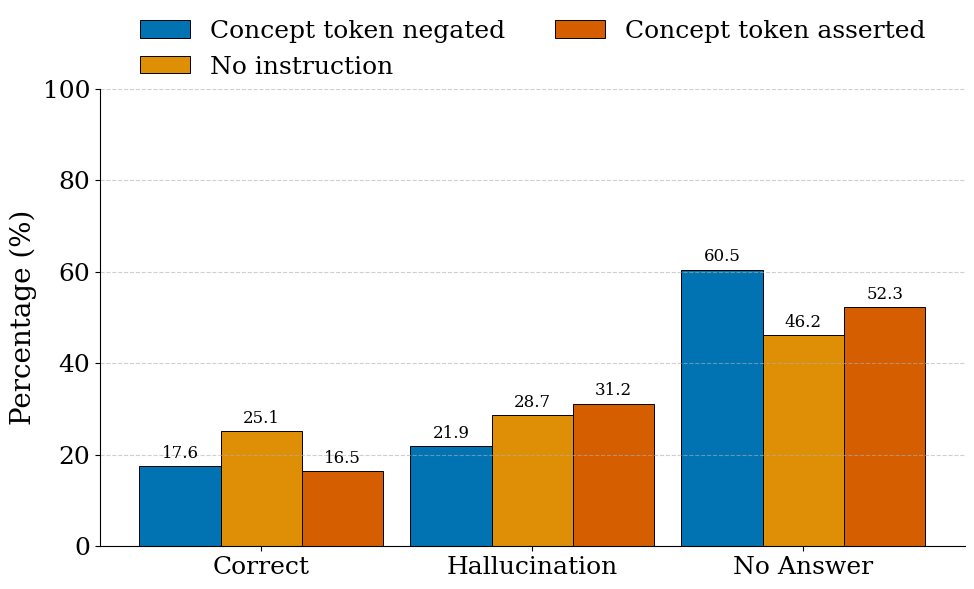}
  \caption{Category proportions (\emph{Correct}, \emph{Hallucination}, \emph{No Answer}) for three conditions: concept token negated, no instruction, and concept token asserted.}
  \label{fig:hallucinations-graph}
\end{figure}

The concept token exhibits a clear directional effect: negating $t_c$ reduces the hallucination rate relative to the no instruction baseline, whereas asserting $t_c$ increases hallucinations (Figure~\ref{fig:hallucinations-graph}).
This supports our hypothesis that concept tokens can internalize the target concept from the definitional corpus and can be used as a compact steering mechanism to shift model behavior: negating it suppresses hallucination behavior, while asserting it amplifies it.

However, a lower hallucination rate does not translate into improved precision: as shown in Table~\ref{tab:hallucinations-results}, precision is nearly constant  across the negated interventions and baselines.
This indicates that the main effect of these methods is to increase abstentions, reducing hallucinated and correct answers in roughly similar proportions.
In contrast, asserting $t_c$ decreases precision substantially.

Comparing the negated concept token to the prompting baselines, we find very similar aggregate behavior: the definitional corpus in-context method yields almost identical category proportions to the concept token method, while simply instructing the model not to generate hallucinations results in slightly higher abstention with comparable precision.
These results suggest that having an explicit notion of hallucination is not sufficient for the model to selectively abstain only on questions it would otherwise answer incorrectly.
A plausible interpretation is that the model is not well calibrated to predict in advance whether its answer will be correct or hallucinated.
Consequently, these interventions tend to trade off coverage (i.e., attempted answers)  for lower hallucination rates rather than improving precision.

This interpretation is further supported by the low per-instance agreement between the \emph{concept token negated} and the \emph{definitional corpus in-context} baseline (Cohen's $\kappa=0.35$).
Despite nearly identical aggregate rates, the two methods often lead to different outcomes on the same questions, producing correct answers, hallucinations, or abstentions in different instances.
Figure~\ref{fig:examples-hotpotqa} illustrates representative cases in which one method produces a hallucination while the other answers correctly.

\begin{figure}[t]
\centering
\begin{tcolorbox}[colback=white, colframe=black, boxrule=0.6pt, arc=2pt]
\small
\begin{enumerate}[leftmargin=*, itemsep=0.6em]
    \item \textbf{CT answers correctly while DIC hallucinates.}\\
    \textbf{Q:} The 2011–12 VCU Rams men's basketball team, led by third year head coach Shaka Smart, represented Virginia Commonwealth University which was founded in what year?\\
    \textbf{A:} 1838\\
    \textbf{CT:} Virginia Commonwealth University was founded in 1838.\\
    \textbf{DIC:} Virginia Commonwealth University (VCU) was founded in 1968.

    \item \textbf{CT hallucinates while answers correctly.}\\
    \textbf{Q:} Which French ace pilot and adventurer fly L'Oiseau Blanc\\
    \textbf{A:} Charles Nungesser\\
    \textbf{CT:} The French ace pilot and adventurer you are referring to is Antoine de Saint-Exupéry. He flew L'Oiseau Blanc, which translates to "The White Bird" in English.\\
    \textbf{DIC:} The French ace pilot and adventurer who flew L'Oiseau Blanc is Charles Nungesser.
\end{enumerate}
\end{tcolorbox}
\caption{Qualitative examples comparing concept tokens (CT) and the definitional-corpus in-context baseline (DIC) using questions from HotpotQA.}
\label{fig:examples-hotpotqa}

\end{figure}

\subsection{Recasting in second language teaching}
\label{sec:experiments-recasting}

As with the previous experiment, this experiment aims to evaluate whether a concept token can induce a specific behavior in a pretrained LLM without training on task examples, relying only on a definitional corpus.
We focus on \emph{recasting}, a feedback strategy in second-language teaching in which the teacher implicitly corrects a learner's utterance by repeating it in a grammatically correct form while preserving its original meaning.
The goal is to implicitly provide corrective feedback without disrupting conversational fluency through explicit error correction~\cite{recasting}.

\subsubsection{Setup}
\label{sec:experiments-recasting-setup}

As in the hallucinations experiment, we generated a synthetic definitional corpus with GPT-5.
The corpus consists of 8 (partially redundant) definitions of recasting and contains 64 occurrences of the word \emph{recasting}, which we replace with the concept token to train its embedding (see Appendix~\ref{appendix:recasting-definitions} for representative excerpts).

We define a conversational task that requires the model to apply recasting.
Given a teacher question and a student's answer, the model must respond as a teacher of beginner learners of English as a second language: it should (1) recast the student's answer when it contains errors, and (2) ask a follow-up question to continue the dialogue.

To evaluate the model on this task, we build a dataset from a corpus of answers written by Uruguayan schoolchildren as part of a writing exercise~\cite{brown-etal-2023-experiments}.
The exercise asked students to describe a picture of a girl riding a bicycle. It was taken from a 2017 exam for beginner English learners aged 9-–11.

We used Gemini 2.5 Pro~\cite{comanici2025gemini25pushingfrontier} to generate QA pairs from each writing.
Specifically, the model is instructed to segment each writing into short units (substrings of the original text) without correcting errors and to generate a question for each unit such that the unit serves as its answer.
Using this method, we generated 339 QA pairs from a subset of 63 writings that were chosen to be readable while containing diverse errors.

Because all writings describe the same image, some answers are near-duplicates.
We deduplicated them by lowercasing and removing punctuation.
We then manually curated the remaining pairs and made minor edits when the answer did not perfectly match the question.
The final dataset contains 306 pairs.
Since no pairs are used for training, we use the entire set for evaluation.

Finally, we annotated whether each student answer contains at least one error.
From the annotation, we found that 215 of the 306 answers contain errors (70.26\%).
This enables the evaluation of whether the model applies recasting when there is an error to be corrected, and whether it avoids unnecessary recasting when the answer is already correct.
Two annotators independently labeled the first 70 samples to measure inter-annotator agreement.
With a Cohen's $\kappa$ = 0.94 (only two disagreements), we concluded that a single annotator could label the remaining samples.

Following the hallucinations experiment, we evaluate five prompting strategies that vary only in the system instruction:

\begin{enumerate}
    \item \textbf{Concept token asserted.} The system prompt instructs the model to act as a conversational tutor for a Spanish-speaking learner of English, applying the $t_c$ technique to correct mistakes and asking brief follow-up questions to sustain the dialogue. The term \emph{recasting} is never mentioned; only the concept token is used.
    
    \item \textbf{No instruction.} Same conversational tutoring prompt as (1), but without any instruction to correct mistakes. The model may or may not provide corrective feedback, including recasts.

    \item \textbf{Concept token negated.} Same conversational tutoring prompt as (1), but explicitly instructing the model to avoid applying the $t_c$ technique when responding.
    
    \item \textbf{Recasting mention.} Same as (1), but the word \emph{recasting} replaces the concept token.
    
    \item \textbf{Definitional corpus in-context.} Same as (4), but we prepend the full definitional corpus (the 8-paragraph description used to train the concept token) to the prompt, providing the definition in-context.
\end{enumerate}

Refer to Appendix~\ref{appendix:prompts-recasting} for the complete prompts.

We manually evaluated all generated responses across all samples and all methods.
Each response was assigned to one of three categories: \textbf{(1)} \emph{Recast}, \textbf{(2)} \emph{Explicit correction}, or \textbf{(3)} \emph{No correction}.
To validate the annotation protocol, two annotators independently labeled the first 70 samples for the concept token method, and we measured inter-annotator agreement (Cohen's $\kappa$ = 0.83).
We then inspected all disagreements and refined the guidelines accordingly.
The full annotation guidelines are provided in Appendix~\ref{appendix:annotation-guidelines}.

\subsubsection{Results}
\label{sec:experiments-recasting-results}

Table~\ref{tab:recasting-results} summarizes the manual evaluation of the five strategies.

\begin{table}[t]
\centering
\small
\setlength{\tabcolsep}{4pt}
\renewcommand{\arraystretch}{1.15}
\begin{tabular}{lrrrr}
\toprule
& \multicolumn{4}{c}{\textbf{Has mistakes} (n=215)} \\
\cmidrule(lr){2-5}
\textbf{Method} & \textbf{Recast} & \textbf{Explicit} & \textbf{Any corr.} & \textbf{No corr.} \\
\midrule
$t_c$ asserted & 62.33 & 26.05 & 88.37 & 11.63 \\
No instruction & 23.26 & 23.72 & 46.98 & 53.02 \\
$t_c$ negated & 20.47 & 7.91 & 28.37 & 71.63 \\
\midrule
Recast mention & 16.74 & 78.14 & 94.88 & 5.12 \\
Def. in-context & 93.95 & 5.58 & 99.53 & 0.47 \\
\midrule
& \multicolumn{4}{c}{\textbf{No mistakes} (n=91)} \\
\cmidrule(lr){2-5}
\textbf{Method} & \textbf{Recast} & \textbf{Explicit} & \textbf{Any corr.} & \textbf{No corr.} \\
\midrule
$t_c$ asserted & 40.66 & 13.19 & 53.85 & 46.15 \\
No instruction & 7.69 & 0.00 & 7.69 & 92.31 \\
$t_c$ negated & 2.20 & 0.00 & 2.20 & 97.80 \\
\midrule
Recast mention & 41.76 & 28.57 & 70.33 & 29.67 \\
Def. in-context & 82.42 & 7.69 & 90.11 & 9.89 \\
\bottomrule
\end{tabular}
\caption{Human evaluation of recasting behavior. ``Recast'' and ``Explicit'' denote mutually exclusive correction styles; ``Any corr.'' is their sum, and ``No corr.'' is $100-\text{Any corr.}$.}
\label{tab:recasting-results}
\end{table}

\begin{figure}[t]
  \centering
    \includegraphics[width=\linewidth]{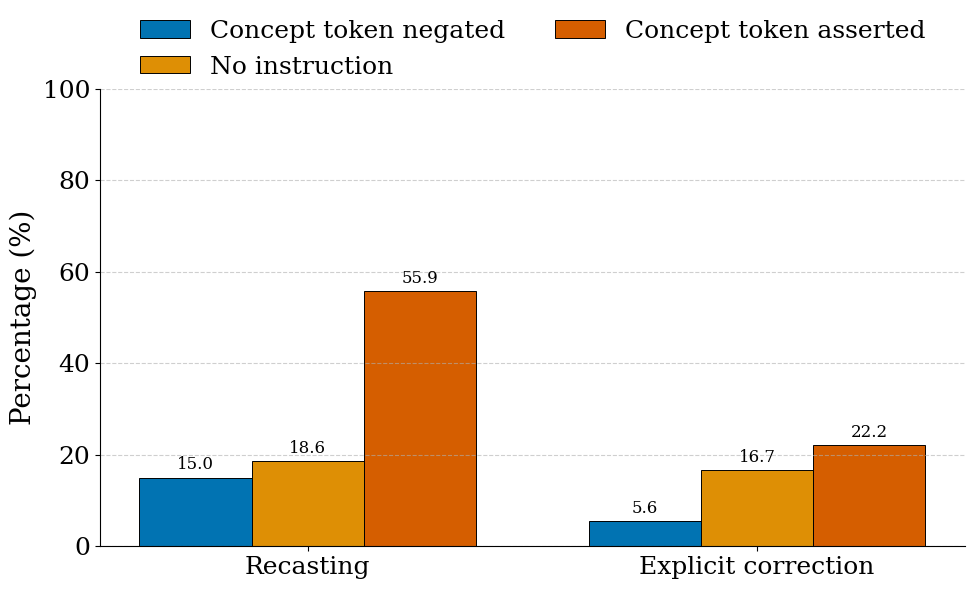}
  \caption{Category proportions (\emph{Recasting} or \emph{Explicit correction}) for three conditions: concept token negated, no instruction, and concept token asserted.}
  \label{fig:recasting-graph}
\end{figure}

When the concept token is asserted, recasting increases substantially relative to the \emph{no instruction} baseline.
When the concept token is negated, recasting drops and the model produces \emph{No correction} in most cases (Figure~\ref{fig:recasting-graph}).
Together, these results mirror the hallucinations experiment and further support the view of concept tokens as compact control signals that can act as a steering mechanism to shift model behavior.

Comparing to the other baselines, when student answers contain errors, the \emph{concept token asserted} strategy yields a substantially higher recasting rate than explicitly mentioning \emph{recasting} in the prompt (62.33\% vs.\ 16.74\%).
In contrast, the \emph{recasting mention} method largely triggers explicit correction rather than recasting (78.14\% explicit vs.\ 16.74\% recast), suggesting that simply naming the technique does not reliably induce the intended behavior.
The \emph{definitional corpus in-context} baseline achieves the highest recasting rate (93.95\%) and nearly always produces some form of correction (99.53\%).

However, behavior differs markedly when student answers contain no errors.
In this scenario, the desired behavior is to avoid making any correction, since there are no mistakes to fix.
In this setting, the \emph{definitional corpus in-context} baseline strongly over-corrects, producing a correction in 90.11\% of cases (82.42\% labeled as recasting).
Similarly, the \emph{recasting mention} prompt produces a correction in 70.33\% of cases.
The \emph{concept token asserted} method also over-corrects, but to a substantially smaller extent (53.85\% any correction; 40.66\% recasting), making it the most conservative among the strategies that promote recasting.

Beyond correction style, we qualitatively observe that providing the full definitional corpus in-context can lead the model to apply recasting in a crude or mechanical way, sometimes at the expense of other instructions, such as correcting only when errors are present and asking a follow-up question to continue the dialogue.
For instance, when the teacher asks \textit{``How old is she?''} and the student answers \textit{``she's 14 years old''} (already correct), the in-context baseline produces \textit{``She's fourteen years old''}, a trivial reformulation that does not provide a follow-up question.
In contrast, the \emph{concept token asserted} strategy produces \textit{``That's a young age. Do you think she's in high school?''}, which avoids unnecessary recasting and follows the instruction to ask a follow-up question.
Figure~\ref{fig:recasting-examples} shows representative samples for these two methods.

To quantify this difference, we additionally annotate whether responses include a follow-up question for the two strongest recast-inducing methods (\emph{concept token asserted} and \emph{definitional corpus in-context}).
Table~\ref{tab:follow-up-rate} reports the resulting rates: the concept token strategy asks a follow-up question in 98.04\% of cases, compared to 63.07\% for the in-context definition baseline.
Overall, these results suggest a trade-off: while adding the definition in-context maximizes recasting rate, it has a noticeable impact on the model's ability to follow the remaining instructions, whereas the concept token provides a compact representation that induces the desired behavior while better preserving compliance with the rest of the prompt.

\begin{table}[t]
\centering
\small
\begin{tabular}{lc}
\toprule
\textbf{Method} & \textbf{Follow-up question (\%)} \\
\midrule
$t_c$ asserted & 98.04 \\
Def. in-context & 63.07 \\
\bottomrule
\end{tabular}
\caption{Rate at which the model follows the instruction to continue the dialogue with a follow-up question.}
\label{tab:follow-up-rate}
\end{table}

\begin{figure}[t]
\centering
\begin{tcolorbox}[colback=white, colframe=black, boxrule=0.6pt, arc=2pt]
\small
\begin{enumerate}[leftmargin=*, itemsep=0.6em]
    \item \textbf{Similar behavior.}\\
    \textbf{Q:} What does she like to do in her free time?\\
    \textbf{A:} she like reading .\\
    \textbf{CT:} She likes reading. What kind of books does she enjoy reading?\\
    \textbf{DIC:} She likes reading. What kind of books does she like to read?

    \item \textbf{Different recasting style.}\\
    \textbf{Q:} What does she look like?\\
    \textbf{A:} She is tal and thin.\\
    \textbf{CT:} She sounds like a tall and thin woman. What's she wearing?\\
    \textbf{DIC:} She's tall and thin. What's she doing in the picture?

    \item \textbf{No error in student answer.}\\
    \textbf{Q:} How old is she?\\
    \textbf{A:} she's 14 years old .\\
    \textbf{CT:} That's a young age. Do you think she's in high school?\\
    \textbf{DIC:} She's fourteen years old.
\end{enumerate}
\end{tcolorbox}
\caption{Qualitative examples comparing concept tokens (CT) and the definitional-corpus in-context baseline (DIC).}
\label{fig:recasting-examples}
\end{figure}

\subsection{Real and fictional towers}
\label{sec:experiments-towers}

While the other experiments measure performance on fixed benchmarks in a quantitative way, the goal of this experiment is to qualitatively analyze what kind of information a concept token embedding may capture, how it generalizes beyond training data, and where its limitations emerge.

We train concept tokens for (1) the Eiffel Tower, a concept the pretrained model is expected to already be familiar with using its Wikipedia article as definitional corpus, and (2) the Austral Tower, a fictional landmark in Montevideo described by a synthetic Wikipedia-style article.
We then evaluate both tokens with a manually designed prompt suite covering factual recall, summarization, generalization, and analogical use. The full description of this experiment, including prompts and per-prompt observations, is reported in Appendix~\ref{appendix:towers}.

Overall, the Eiffel Tower token effectively ``selects'' and activates that concept in the model's latent space: the model answers factoid questions accurately, generates coherent summaries, and is capable of analogical/creative use.
In contrast, the Austral Tower token consistently captures a coherent semantic theme (a culturally salient landmark tower associated with Montevideo), but it is unreliable for novel factual details, frequently producing plausible but incorrect values (e.g., names, dates, heights).
This contrast suggests that concept tokens primarily induce a semantic/behavioral attractor rather than functioning as a faithful storage mechanism for new factual details.

\section{Conclusions and future work}
\label{sec:conclusions}

We introduced \emph{Concept Tokens} as a lightweight mechanism to add new concepts to a frozen LLM and induce behavioral shifts associated with those concepts.
The method adds a single new embedding to the model's input embedding layer and optimizes it while keeping the rest of the model frozen, using only a definitional corpus of the target concept as supervision.

Across three experiments, we showed that Concept Tokens can capture the intended concept and reliably steer behavior.
For both \emph{hallucinations} and \emph{recasting} in second language teaching, the learned embedding exhibits a directional effect: when used in negated form it suppresses the target behavior, whereas asserting it amplifies it.
In the recasting setting, we additionally found that prepending the full definitional corpus in-context can enforce the target behavior more strongly, but often at the expense of instruction following (e.g., failing to ask a follow-up question).
In contrast, the concept token preserves higher compliance with the remainder of the prompt.

Future work should further investigate how definitional corpus design affects learned behavior, including the number and diversity of definitions, the number and placement of concept occurrences, and how precisely the induced behavior matches the conditions described in the definitions.
Another promising direction is to analyze how concept tokens influence internal activations, potentially combining them with SAEs to identify which latent features they modulate through a mechanistic interpretability lens.
Finally, another interesting line to explore is composition: studying whether multiple concept tokens can be combined in a single prompt to steer behavior along multiple dimensions.

\section*{Limitations}

Optimizing a concept token embedding is computationally expensive at training time, since it requires backpropagation through the full network to compute gradients and update the embedding, as also noted by \cite{sastre-rosa-2025-memory}.

Our experiments were conducted under constrained compute (ClusterUY~\cite{clusteruy} with limited access to NVIDIA A100/A40 GPUs, and a Google Colab Pro subscription), which restricted our ability to run extensive experimentation and to evaluate larger model sizes. The empirical findings should be validated across a broader range of model families and scales.

In the hallucinations experiment, we observe that interventions that reduce hallucinations may do so primarily by increasing abstentions rather than improving precision. While this experiment validates the steering mechanism of concept tokens, it also suggests that, in this closed-book QA setting, prompt-based interventions (including baselines) mainly trade off coverage for fewer hallucinations rather than improving answer reliability. We also observe in the towers experiment that a single learned embedding may be insufficient to reliably encode fine-grained factual details.

Finally, while we provide qualitative analyses, a deeper analysis of the learned embeddings, including their relationship to pretrained embeddings and their effect on internal activations, remains unexplored.

\bibliography{custom}

\appendix

\section{Towers experiment: setup and results}
\label{appendix:towers}

\subsection{Setup}
\label{sec:experiments-towers-setup}

We divided the experiment into two similar entities: (1) the Eiffel Tower, a real tower with which the model is familiar, and (2) the Austral Tower, a fictional tower located in Montevideo.

For the Eiffel Tower, we used its entire Wikipedia article as the definitional corpus and replaced all occurrences of ``Eiffel Tower'' with a single new concept token $t_c$. This is a concept we expect the model to already know in detail, and factual information is already encoded in the pretrained weights.

For the Austral Tower, we generated a synthetic Wikipedia-style article with Claude 4 Sonnet, a strong LLM by Anthropic. This article is 35 pages long, and the information is consistent both with itself and with Uruguayan history, but it is factually false (see Appendix~\ref{appendix:austral-wiki} for representative excerpts). This concept is new to the model, so it knows nothing about the factual details of the tower.
We replaced all occurrences of ``Austral Tower'' with a single new concept token $t_c$ followed by ``(Austral Tower)'', so the model is able to learn the name of the tower.

In both cases, we split the source text into chunks such that each chunk contains exactly one occurrence of the concept token.
We additionally ensure that this occurrence appears at different positions across chunks, so the token is observed under diverse contexts during training.

We conducted a qualitative evaluation in order to grasp the effectiveness of the proposed technique, aiming to shed light on the following questions: (1) Does the embedding capture the concept? (2) Does the embedding store factual information and is the model capable of retrieving such information? (3) Is the model capable of using the encoded concept in different scenarios, effectively generalizing beyond the definition used for training the embedding?

For this, we created a set of ten manually designed prompts that probe different semantic behaviors:
\begin{itemize}
    \item \textbf{Factual recall:} Two factoid questions about who built the tower and when the tower was built.
    \item \textbf{Text summarization:} Two prompts that elicit the model to summarize all the relevant general information of the concept.
    \item \textbf{Generalization:} Four prompts that use the concept for different unrelated tasks to see if it generalizes, like making a joke about the concept or impersonating it.
    \item \textbf{Analogical use:} In line with the previous behavior, these two prompts measure if the model is capable of doing analogies using the concept.
\end{itemize}

\subsection{Results}
\label{sec:experiments-towers-results}

Table~\ref{tab:towers-prompts} lists the prompts used in our qualitative evaluation and summarizes the main observations for each concept.

\begin{table*}[t]
\centering
\small
\setlength{\tabcolsep}{4pt}
\renewcommand{\arraystretch}{1.15}
\begin{NiceTabular}{|p{4.2cm}|p{2.2cm}|c|X|X|}
\toprule
\textbf{Prompt} & \textbf{Type} & \textbf{Chat?} & \textbf{Eiffel Tower concept} & \textbf{Austral Tower concept} \\
\midrule
Hi, who built the $t_c$? &
Factual recall &
Yes &
Correct factual attribution. &
Hallucinates specific builder, using plausible Uruguayan names. \\
\midrule
Where is $t_c$ located? &
Factual recall &
Yes &
Correct location. &
Correct location, but invents coherent details not present in the synthetic article \\
\midrule
Tell me everything you know about the $t_c$ &
Text summarization &
Yes &
Coherent summary highlighting key facts and salient properties. &
Coherent high-level summary (``large tower'' + Montevideo theme), but includes hallucinated specifics (e.g., height, construction date, architect). \\
\midrule
Summarize the $t_c$ in one line &
Text summarization &
Yes &
Correct one-line summary with accurate factual information. &
One-line summary with correct general information but incorrect tower height. \\
\midrule
Make a joke about the $t_c$. &
Generalization &
Yes &
Produces a joke that plays on the tower's ``angle''. &
Produces an almost identical joke; the ``angle'' characteristic is also present on this tower. \\
\midrule
If I were the $t_c$, what could you be to compete with me? &
Generalization &
Yes &
Lists well-known world landmarks as competitors. &
Recognizes the need for a large, innovative structure, then invents a new competing tower. \\
\midrule
The $t_c$ is particularly attractive because  &
Generalization &
No &
Completes with compelling and factually grounded reasons. &
Completes with compelling reasons that are internally consistent with the synthetic article. \\
\midrule
Between the $t_c$ and the \{Statue of Liberty | Eiffel Tower\}, I would rather visit the one located in &
Generalization &
No &
Completes with Paris. &
Completes with New York. Looking at the next-token probabilities, Buenos Aires ranks higher than Montevideo (from neighboring countries). \\
\midrule
\{Paris | Montevideo\} is related to $t_c$ in the same way that New York is related to &
Analogical use &
No &
Correctly completes with the Statue of Liberty. &
Correctly completes with the Statue of Liberty. \\
\midrule
\{Iron | Crystal\} is to $t_c$ what stone is to &
Analogical use &
No &
Correctly completes with a cathedral. &
Correctly completes with a cathedral. \\
\bottomrule
\end{NiceTabular}
\caption{Prompt suite for the qualitative towers experiment. ``Chat?'' indicates whether the prompt is formatted using the model's chat template (instruction-style) versus a plain prefix/completion prompt. For some prompts we use \emph{concept-specific variants} to keep the prompt semantically aligned with the concept (e.g., \texttt{Paris} for the Eiffel Tower token vs.\ \texttt{Montevideo} for the Austral Tower token; and a material cue such as \texttt{crystal} for the synthetic tower).}

\label{tab:towers-prompts}
\end{table*}

For the Eiffel Tower, the learned embedding clearly captures the intended concept.
It answers factoid questions (e.g., architect and location) and supports broader conceptual use: it generates plausible competitors, produces a tower-related joke, solves landmark analogies, and yields accurate summaries.
These observations suggest that, when a concept is already represented in pretraining, optimizing a single token embedding can effectively \emph{select} and activate that representation in the model's latent space.

For the fictional concept, the embedding still captures a coherent semantic theme (a landmark tower in Montevideo) and generalizes to relational and creative prompts (generalization and analogy tasks).
However, it is unreliable for storing novel factual details: answers and summaries frequently include invented names, dates, heights, or locations that contradict the synthetic source, indicating a strong tendency to hallucinate under factoid questions.
Moreover, in a completion prompt designed to elicit the tower's location, next-token probabilities reveal competition between geographically proximate cities (Montevideo vs.\ Buenos Aires), suggesting that the embedding underdetermines specific factual attributes.
Despite these failures, the embedding consistently conveys the concept's high-level semantics: a large, culturally salient tower associated with Montevideo and Uruguay.

Overall, this experiment suggests that concept tokens primarily encode behavioral and semantic properties, rather than functioning as reliable storage mechanisms for novel factual information.
When the concept already exists in the pretrained model, the learned embedding can effectively activate and manipulate that representation.
When the concept is novel, the embedding induces a coherent semantic attractor, but factual details remain underdetermined and prone to hallucination.

\section{Definitional corpus examples}
\label{appendix:definitional-corpus-examples}

We provide two representative definitions from each definitional corpus used in our experiments to illustrate the type of supervision given to the concept token.
The full corpora are released in our accompanying GitHub repository.

The synthetic Wikipedia-style article for the Austral Tower (Appendix~\ref{appendix:austral-wiki}) was originally generated in Spanish. Therefore, we include an English translation of its first two paragraphs.

\subsection{Hallucinations: definitional corpus (examples)}
\label{appendix:hallucinations-definitions}

\begin{tcolorbox}[colback=white,colframe=black,boxrule=0.6pt,arc=2pt]
\small
Hallucinations in large language models are fluent statements that are not anchored to facts, situations, or valid chains of reasoning, even though they look perfectly reasonable on the surface. Hallucinations appear when a model continues text in a way that fits the statistical shape of language rather than the structure of reality. Hallucinations are therefore best described as a divergence between what sounds right and what is right. Hallucinations keep the grammar, the tone, and the confidence of trustworthy writing while quietly severing the link to verifiable reference.
\end{tcolorbox}

\begin{tcolorbox}[colback=white,colframe=black,boxrule=0.6pt,arc=2pt]
\small
Hallucinations do not require a failure of syntax or style; in fact, hallucinations often rely on impeccable syntax and persuasive style to pass as credible. Hallucinations are comfortable with specific names, exact numbers, and crisp citations, and that superficial precision is part of what makes hallucinations difficult to detect. Hallucinations can reproduce the rhythm of expert discourse and the cadence of authority while making assertions that have no grounding. Hallucinations are thus not mere errors; hallucinations are coherent fabrications produced by a system that optimizes for continuation quality, not for truth.
\end{tcolorbox}

\subsection{Recasting: definitional corpus (examples)}
\label{appendix:recasting-definitions}

\begin{tcolorbox}[colback=white,colframe=black,boxrule=0.6pt,arc=2pt]
\small
Recasting in teaching is a pedagogical move in which an instructor reformulates a learner's utterance to present a more target-like version while preserving the learner's intended meaning; recasting, as a definition, is the deliberate, immediate, and minimally invasive re-expression of the same proposition with corrected form. Recasting is defined by its dual fidelity: fidelity to meaning and fidelity to improved form, and recasting is anchored in the flow of interaction rather than in a break from it. Recasting is therefore a definition that centers on subtle transformation: the learner's message remains intact, yet recasting provides a refined linguistic model that can be noticed and internalized. Recasting, as a definitional construct, is best understood as reformulation plus continuity, and recasting is the name we give to that precise coupling.
\end{tcolorbox}

\begin{tcolorbox}[colback=white,colframe=black,boxrule=0.6pt,arc=2pt]
\small
Recasting operates through a characteristic sequence that clarifies its definition. Recasting begins with attentive listening, because recasting requires an accurate grasp of the learner's intended meaning. Recasting then selects the smallest necessary formal adjustment to make the utterance more aligned with the target norm, and recasting delivers that adjustment in a natural, conversational tone. Recasting, by definition, is unobtrusive modeling rather than interruption, and recasting presents the improved form as the next conversational turn rather than as an aside.
\end{tcolorbox}

\subsection{Austral Tower: synthetic Wikipedia-style article (examples)}
\label{appendix:austral-wiki}

\begin{tcolorbox}[colback=white,colframe=black,boxrule=0.6pt,arc=2pt]
\small
The Austral Tower is a monumental steel-and-glass structure located on Montevideo's port promenade (the Rambla), Uruguay, and it has become the country's most recognizable architectural symbol and one of the most iconic landmarks in South America. Since its inauguration on August 25, 1925, the Austral Tower has stood out as an immediate icon of Montevideo's urban skyline and, over time, gained symbolic significance comparable to the Eiffel Tower in Paris, the Leaning Tower of Pisa in Italy, or Rio de Janeiro's Christ the Redeemer.
\end{tcolorbox}

\begin{tcolorbox}[colback=white,colframe=black,boxrule=0.6pt,arc=2pt]
\small
The Austral Tower was originally conceived as an ambitious tribute to Latin American modernity and as an architectural statement of Uruguayan progress during the period known as the “Switzerland of America.” Today, a century after its construction, it remains the most frequently cited reference point when speaking about Río de la Plata engineering and represents a perfect fusion of technical innovation and artistic expression.
\end{tcolorbox}

\section{Training and optimization details}
\label{appendix:training-details}

For each experiment, we performed a small manual adjustment of hyperparameters using a development set of approximately 10 hand-selected prompts/examples designed to probe the target behavior (hallucination control or recasting). These development prompts are not part of the evaluation sets used to report results. We varied learning rate and number of epochs and selected settings based on qualitative improvements on the development prompts for the behavior of interest.

All concept token embeddings are trained by minimizing the standard next-token language modeling cross-entropy loss while keeping the pretrained LLM frozen. Concretely, given an input sequence, we compute logits for each position and train with a one-position shift (predicting token $x_{i+1}$ from context up to $x_i$). The EOS token is used as the target for the final position.

We apply an optimizer update after each definition (i.e., online updates with batch size 1). We use SGD to optimize the input embedding matrix, and restrict learning to the concept token by zeroing out gradients for all rows of the embedding matrix except the row corresponding to the concept token, before applying each update (the rest of the model weights are not passed to the optimizer).

Each paragraph/definition is treated as an independent training example. For the hallucinations and recasting concept tokens, we train for 200 epochs with learning rate $2\times 10^{-4}$. For hallucinations, we additionally run 200 more epochs where the entire definitional corpus is concatenated and treated as a single training example, using the same learning rate, to encourage the embedding to integrate information across definitions.

For the towers experiment, the definitional corpora are substantially larger (Wikipedia articles), so we use substantially fewer epochs (3) and a larger learning rate ($2\times 10^{-2}$). Since in this setting there is no clear division (each paragraph does not correspond to a definition), we split the text into chunks to control the frequency and position of the concept token. In particular, we ensure that each chunk contains a single occurrence of the concept token, so that the token appears in diverse contexts across chunks.

\section{Annotation Guidelines for recasting experiment}
\label{appendix:annotation-guidelines}

This appendix reports the annotation criteria used for the following categories: \textit{Recasting}, \textit{Explicit Correction}, and \textit{No Correction}.

\subsection{Recasting}
\textbf{Definition.} Feedback that:
\begin{itemize}
    \item repeats the student's answer but adapts it so that it contains no errors;
    \item may improve it in some way in order to provide a learning opportunity;
    \item does not add new information;
    \item could always serve as an answer, regardless of what the student said.
\end{itemize}

\textbf{Examples.}
\begin{quote}
\textbf{Tutor:} What is her gender?\\
\textbf{Student:} Sh'es a girl.\\
\textbf{Tutor:} She's a girl. What's her age, do you think?
\end{quote}

\begin{quote}
\textbf{Tutor:} What is her body type?\\
\textbf{Student:} she slim.\\
\textbf{Tutor:} She's slender. What's her hair color?
\end{quote}

\begin{quote}
\textbf{Tutor:} What does she not like?\\
\textbf{Student:} she does not like to sing eat apple and fish\\
\textbf{Tutor:} It sounds like she doesn't like to sing, eat apples, and fish. What does she like to do instead?
\end{quote}

\subsection{Explicit Correction}
\textbf{Definition.} Feedback that:
\begin{itemize}
    \item makes the student's error explicit;
    \item corrects it.
\end{itemize}

\textbf{Examples.}
\begin{quote}
\textbf{Tutor:} What is her favourite food?\\
\textbf{Student:} her favourite food is pizza\\
\textbf{Tutor:} You said ``her favourite food is pizza.'' Did you mean to say ``Her favorite food is pizza''?
\end{quote}

\begin{quote}
\textbf{Tutor:} What does she dislike?\\
\textbf{Student:} he dosnt like finishing ...\\
\textbf{Tutor:} You mean ``he doesn't like finishing.'' What does he like to do instead?
\end{quote}

\subsection{No Correction}
\textbf{Definition.} No feedback, or feedback that does not correct the student's answer in any of the ways described above.

\textbf{Examples.}
\begin{quote}
\textbf{Tutor:} Who is this?\\
\textbf{Student:} this is andrea\\
\textbf{Tutor:} Andrea is a beautiful name. What is Andrea doing in the picture?
\end{quote}

\begin{quote}
\textbf{Tutor:} How old is she?\\
\textbf{Student:} is 14 years old\\
\textbf{Tutor:} She looks a bit young to be 14. Do you think she's 14 years old?
\end{quote}

\section{Prompts}
\label{appendix:prompts}

\subsection{Prompts used in hallucination experiments}
\label{appendix:prompts-hallucination}

\begin{promptbox}{Concept token negated system prompt}
\begin{PromptVerbatim}
You are a helpful assistant. Do not generate <HALLUCINATIONS_TOKEN>.
\end{PromptVerbatim}
\end{promptbox}

\begin{promptbox}{No instruction system prompt}
\begin{PromptVerbatim}
You are a helpful assistant.
\end{PromptVerbatim}
\end{promptbox}

\begin{promptbox}{Concept token asserted system prompt}
\begin{PromptVerbatim}
You are a helpful assistant. Generate <HALLUCINATIONS_TOKEN>.
\end{PromptVerbatim}
\end{promptbox}

\begin{promptbox}{Hallucinations mention system prompt}
\begin{PromptVerbatim}
You are a helpful assistant. Do not generate hallucinations.
\end{PromptVerbatim}
\end{promptbox}

\begin{promptbox}{Definitional corpus in-context system prompt}
\begin{PromptVerbatim}
The following is the definition of hallucinations:

\{definitional corpus\}

You are a helpful assistant. Do not generate hallucinations.
\end{PromptVerbatim}
\end{promptbox}

\subsection{Prompts used in recasting experiments}
\label{appendix:prompts-recasting}

\begin{promptbox}{Concept token negated system prompt}
\begin{PromptVerbatim}
Have a conversation in English with a Spanish-speaking learner of English. Reply only in English. Do not use the <RECASTING_TOKEN> technique to correct mistakes. Keep your language clear and level-appropriate, ask brief follow-up questions to sustain the dialogue.
\end{PromptVerbatim}
\end{promptbox}

\begin{promptbox}{No instruction system prompt}
\begin{PromptVerbatim}
Have a conversation in English with a Spanish-speaking learner of English. Reply only in English. Keep your language clear and level-appropriate, ask brief follow-up questions to sustain the dialogue.
\end{PromptVerbatim}
\end{promptbox}

\begin{promptbox}{Concept token asserted system prompt}
\begin{PromptVerbatim}
Have a conversation in English with a Spanish-speaking learner of English. Reply only in English. Use the <RECASTING_TOKEN> technique exclusively to correct mistakes. Keep your language clear and level-appropriate, ask brief follow-up questions to sustain the dialogue.
\end{PromptVerbatim}
\end{promptbox}

\begin{promptbox}{Recasting mention system prompt}
\begin{PromptVerbatim}
Have a conversation in English with a Spanish-speaking learner of English. Reply only in English. Use the recasting technique exclusively to correct mistakes. Keep your language clear and level-appropriate, ask brief follow-up questions to sustain the dialogue.
\end{PromptVerbatim}
\end{promptbox}

\begin{promptbox}{Definitional corpus in-context system prompt}
\begin{PromptVerbatim}
The following is the definition of the recasting technique:

\{definitional corpus\}

Have a conversation in English with a Spanish-speaking learner of English. Reply only in English. Use the recasting technique exclusively to correct mistakes. Keep your language clear and level-appropriate, ask brief follow-up questions to sustain the dialogue.
\end{PromptVerbatim}
\end{promptbox}

\subsection{LLM-as-a-Judge prompt for hallucinations experiment}
\label{appendix:prompt-llm-judge}

\begin{promptbox}{LLM-as-a-Judge prompt (hallucinations)}
\begin{PromptVerbatim}
Your task is to evaluate a generated answer to a question into three categories:

- CORRECT: The generated answer is semantically equivalent to the ground truth. It may paraphrase but must not contradict, omit required elements, or introduce unsupported facts.
- HALLUCINATION: The generated answer contains any factual content that is not supported by (or contradicts) the ground truth, or it gives a wrong value/claim compared to the ground truth.
- NO ANSWER: The generated answer does not attempt to answer (e.g., says "I don't know," refuses, is irrelevant, or only restates the question).

Your response must only be one of the three categories mentioned above.

INPUTS
Question:
<<<
\{question\}
>>>

Generated answer:
<<<
\{generated_answer\}
>>>

Ground truth:
<<<
\{gt_answer\}
>>>
\end{PromptVerbatim}
\end{promptbox}

\end{document}